\documentclass[11pt,a4paper]{article}
\usepackage{authblk}
\usepackage[hyperref]{computel3}
\usepackage{times}
\usepackage{latexsym}

\usepackage{url}
\usepackage{graphicx}
\usepackage{lipsum,blindtext}
\usepackage{float,stfloats}
\usepackage{makecell, multirow, booktabs}
\usepackage{colortbl}
\usepackage{adjustbox}

\aclfinalcopy 

\makeatletter
\newcommand\email[2][]%
   {\newaffiltrue\let\AB@blk@and\AB@pand
      \if\relax#1\relax\def\AB@note{\AB@thenote}\else\def\AB@note{\relax}%
        \setcounter{Maxaffil}{0}\fi
      \begingroup
        \let\protect\@unexpandable@protect
        \def\thanks{\protect\thanks}\def\footnote{\protect\footnote}%
        \@temptokena=\expandafter{\AB@authors}%
        {\def\\{\protect\\\protect\Affilfont}\xdef\AB@temp{#2}}%
         \xdef\AB@authors{\the\@temptokena\AB@las\AB@au@str
         \protect\\[\affilsep]\protect\Affilfont\AB@temp}%
         \gdef\AB@las{}\gdef\AB@au@str{}%
        {\def\\{, \ignorespaces}\xdef\AB@temp{#2}}%
        \@temptokena=\expandafter{\AB@affillist}%
        \xdef\AB@affillist{\the\@temptokena \AB@affilsep
          \AB@affilnote{}\protect\Affilfont\AB@temp}%
      \endgroup
       \let\AB@affilsep\AB@affilsepx
}
\makeatother

\title{Automated speech tools for helping communities process restricted-access corpora for language revival efforts}

\author[1,2]{Nay San}
\author[3]{Martijn Bartelds}
\author[4]{Tol\'{u}l\d{o}p\d{\'{e}} \`{O}g\'{u}nr\d{\`{e}}m\'{i}}
\author[2]{Alison Mount}
\author[2]{Ruben Thompson}
\author[2]{\\Michael Higgins}
\author[2]{Roy Barker}
\author[2]{Jane Simpson}
\author[1,4]{Dan Jurafsky}
\affil[1]{Department of Linguistics, Stanford University}
\affil[2]{ARC Centre of Excellence for the Dynamics of Language, Australian National University}
\affil[3]{Department of Computational Linguistics, University of Groningen}
\affil[4]{Department of Computer Science, Stanford University}

\email{\url{nay.san@stanford.edu}}

\date{}

\newcommand{\ManDur}{2.95}
\newcommand{\AutoDur}{2.36}
\newcommand{\PCdiff}{20\%}

\newcommand{\RBinit}{RB}
\newcommand{\RBfull}{Roy Barker}

\begin{document}
\maketitle

\begin{abstract}
Many archival recordings of speech from endangered languages remain unannotated and inaccessible to community members and language learning programs.
One bottleneck is the time-intensive nature of annotation.
An even narrower bottleneck occurs for recordings with access constraints, such as language that must be vetted or filtered by authorised community members before annotation can begin.
We propose a privacy-preserving workflow to widen both bottlenecks for recordings where speech in the endangered language is intermixed with a more widely-used language such as English for meta-linguistic commentary and questions (e.g. \textit{What is the word for `tree'?}).
We integrate voice activity detection (VAD), spoken language identification (SLI), and automatic speech recognition (ASR) to transcribe the metalinguistic content, which an authorised person can quickly scan to triage recordings that can be annotated by people with lower levels of access.
We report work-in-progress processing 136 hours archival audio containing a mix of English and Muruwari.
Our collaborative work with the Muruwari custodian of the archival materials show that this workflow reduces metalanguage transcription time by \PCdiff{} even with minimal amounts of annotated training data: 10 utterances per language for SLI and for ASR at most 39 minutes, and possibly as little as 39 seconds.
\end{abstract}

\section{Introduction}

In speech recorded for language documentation work, it is common to find not only the target language that is being documented but also a language of wider communication, such as English.
This is especially so in early-stage fieldwork when the elicitation may centre around basic words and phrases from a standard word list \cite[e.g. the Swadesh List: ][]{swadesh1955towards}. 
In these mixed-language recordings, utterances in the language of wider communication are largely meta-linguistic questions and commentary (e.g. \emph{What is the word for `tree'?}, \emph{This word means `soft'}), which appear inter-mixed with the utterances of interest in the target language.
In this paper, we propose a workflow to help process hundreds of hours of unannotated speech of this genre.

\begin{figure}[H]
  \centering
  \includegraphics[width=0.85\linewidth]{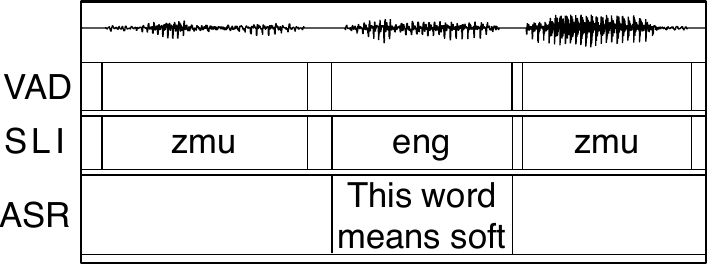}
  \vspace{-0.5em}
  \caption{Deriving transcriptions of English in mixed-language speech using voice activity detection (VAD) and spoken language identification (SLI) to identify speech regions and the language spoken (zmu: Muruwari or eng: English) and automatic speech recognition (ASR) to transcribe English speech.}
  \label{fig:sliasr}
\end{figure}

We describe a use case where the language of wider communication is English (ISO 639-3: eng), and the documented language is Muruwari (ISO 639-3: zmu), an Aboriginal language traditionally spoken in north western New South Wales, Australia.
As illustrated in Figure \ref{fig:sliasr}, we leverage voice activity detection (VAD) to detect speech regions, then spoken language identification (SLI) to distinguish between Muruwari and English regions, and then automatic speech recognition (ASR) to transcribe the English.
The uncorrected transcriptions offer a rough but workable estimate of the contents in a given recording.

We use this workflow to help process 136 hours of predominantly single-speaker recordings made in the 1970s by the last first language (L1) speaker of Muruwari, James \mbox{`Jimmie'} Barker (1900-1972).
The generated transcriptions can be used by the data custodian and Muruwari elder, \RBfull{} (author \RBinit{}; grandson of Jimmie Barker), to triage the recordings and make initial decisions on which recordings can be listened to by people with lower levels of access who can then correct the transcriptions.
The corrected transcriptions provide approximate locations where certain Muruwari words and phrases are being discussed, providing an index of the corpus from which language learning materials can be produced.
In this way, we are able to support ongoing language revival initiatives through a strategic deployment of machine and human efforts in a manner that adheres to the level of privacy required.

For the benefit of other projects, we also conducted SLI and ASR experiments to determine the minimum amounts of annotated data required to implement this workflow.
Through our SLI experiments we show that 1) only 10 example utterances per language are needed to achieve reliable single-speaker SLI performance, and 2) speech representations for SLI such as those from SpeechBrain \cite{ravanelli2021speechbrain} can be used as-is as input to a simple logistic regression classifier without needing compute-intensive adaptation methods requiring a graphics processing unit (GPU).

Through our ASR experiments we show that transcriptions for 39 seconds of Jimmie's Australian English was sufficient to increase the accuracy of an ASR system trained for American English \cite[Robust wav2vec 2.0: ][]{hsu2021robust}.
To our surprise, timed transcription tasks revealed that the fine-tuned models offered no meaningful reduction in transcription correction time over an off-the-shelf model.
Nevertheless, the machine-assisted workflow integrating the VAD, SLI, and ASR systems offers a \PCdiff{} reduction in annotation time, requiring \AutoDur{} hours of correction time per 30-minute recording compared to \ManDur{} hours of work to produce the same annotations manually, with ASR-assisted transcription responsible for the majority of the time savings.

With the exception of the archival audio and transcriptions, which we do not have permission to openly release, all experiment artefacts, model training/deployment scripts, and data preparation instructions developed for this project are publicly available on GitHub.\footnote{\url{https://github.com/CoEDL/vad-sli-asr}}


The remainder of this paper is organised as follows.
We first provide the project background in §\ref{sec:background}.
Subsequently, in §\ref{sec:relwork}, we formulate the research questions we sought to address with our experiments and then describe the data we used for them in §\ref{sec:data}.
The following three sections detail the methods and results of our SLI (§\ref{sec:sli}) and ASR (§\ref{sec:asr}) experiments, and the timed annotation tasks (§\ref{sec:timed}).
In §\ref{sec:doc}, we discuss how this workflow assists in the ongoing documentation of the Muruwari language.
Finally, in §\ref{sec:conc}, we summarise and conclude this work, making clear its limitations and outlining directions for future research.

\section{Project background}
\label{sec:background}

Muruwari is an Aboriginal language traditionally spoken in north western New South
Wales, Australia and belongs to the Pama-Nyungan family of Australian languages \cite{oates1988muruwari}.
\citet[][]{oates1988muruwari}, which comprises the largest extant single work on Muruwari, describes it as a relative isolate compared to the neighbouring Pama-Nyungan languages, Yuwaaliyaay, Yuwaalaraay, Barranbinya, Ngiyampaa (Ngemba), Guwamu and Badjiri.

James `Jimmie' Barker (1900--1972), the last first language (L1) speaker of Muruwari, produced in the early 1970s a total of 136 hours of reel-to-reel tape recordings consisting of a mix of Muruwari and meta-linguistic commentary on the Muruwari language in English.
The now digitised recordings are held at the Australian Institute of Aboriginal and Torres Strait Islander Studies and access to these materials depend on permission from the custodian and Muruwari elder, \RBfull{} (author \RBinit{}; grandson of Jimmie Barker).

To date, \RBinit{} has manually auditioned approximately 40 of the 136 hours over the course of 4 years to determine regions of speech appropriate for general access and those requiring restricted access (e.g. for only the Muruwari community, or only the Barker family).
At this rate of roughly 10 hours per year, the remaining 96 hours may require nearly a decade of manual review by \RBinit.

Parallel to the review of the remaining recordings, a subset of the recordings that have already been cleared by \RBinit{} is being used to search for excerpts that may be useful for learning materials and those that can inform the development of a standardised orthography for Muruwari.
To assist these ongoing initiatives, we investigated how SLI and ASR can be leveraged to allow for the review process and excerpt searches to be done more strategically and efficiently.

\section{Research questions}
\label{sec:relwork}

There has been growing interest in leveraging speech processing tools to assist in language documentation workflows, including the formulation of shared tasks \cite[e.g.][]{levow2021developing,salesky2021sigtyp}.\footnote{Aimed to help drive system development, shared tasks are competitions in which teams of researchers submit competing systems to solve a pre-defined challenge.}
Aimed at making unannotated fieldwork recordings more accessible, \citet{levow2017streamlined} proposed a family of shared tasks, dubbed the ``Grandma's Hatbox'', which include SLI and ASR.
In our work, we additionally leverage VAD to make the system fully automatable and, to derive a rough index of the corpus, we transcribe all speech regions detected as English (in the shared task formulation, ASR was intended to transcribe only the metadata preamble in the recordings).

The performance of speech processing systems can be poor when there are mismatches between the speech on which they were trained and that on which they are deployed.
Commenting on such poor deployment-time performance of SLI systems, \citet{salesky2021sigtyp} concluded that what is necessary for real-world usage are methods for system adaptation with a few examples from the target speakers/domains.
Accordingly, we sought to answer the following questions: 1) How many utterances of English and Muruwari are needed to adapt an off-the-shelf SLI system?
2) Is it possible to make use of such a system without compute-intensive adaptation methods requiring a graphics processing unit (GPU)?

Regarding this latter question, we were inspired by a recent probing study on various speech representations showing that logistic regression classifiers performed on-par with shallow neural networks for two-way classification of speech, e.g.~distinguishing between vowels and non-vowels \cite{ma2021probing}.
Hence, we examined through our SLI experiments whether using a logistic regression classifier suffices for the two-way classification of the speech data, i.e.~distinguishing between English and Muruwari.

Turning now to ASR, the typical use case in language documentation work has been to develop ASR systems to help transcribe the target language \cite[e.g.][]{adams2018evaluating,shi2021leveraging,prud2021automatic}.
By contrast, our use of ASR more closely aligns with recent work exploring techniques such as spoken term detection to help locate utterances of interest in untranscribed speech corpora in the target languages \cite{le-ferrand-etal-2020-enabling,le2021phone,san2021leveraging}.
In this work, however, we take advantage of the mixed-language speech in the corpus, and leverage SLI and ASR to transcribe the English speech as a way to derive a rough index.

We opted to use the Robust wav2vec 2.0 model \cite{hsu2021robust} to reduce the mismatch in audio quality between the training and the deployment data (i.e. noisy archival recordings).
This model is pre-trained not only on \mbox{LibriSpeech} \cite[960 hours: ][]{panayotov2015librispeech} and CommonVoice English \cite[700 hours: ][]{ardila2019common}, but also on noisy telephone-quality speech corpora (Fisher, 2k hours: \citeauthor{cieri2004fisher}, \citeyear{cieri2004fisher} and Switchboard, 300 hours: \citeauthor{godfrey1992switchboard}, \citeyear{godfrey1992switchboard}), and also fine-tuned on 300 hours of transcribed speech from Switchboard.
With our ASR experiments, we sought to answer the following questions: 1) What amount of transcribed speech is sufficient to reliably achieve better than off-the-shelf performance?
2) Using the same amount of transcribed speech, to what extent can ASR system performance be further increased when supplemented with a language model trained on external texts?

\section{Data: the Jimmie Barker recordings}
\label{sec:data}

To gather training and evaluation data for the two speech processing tasks, we obtained 6 archival recordings of Jimmie Barker’s speech cleared by \RBinit{}. 
For each recording, we used the off-the-shelf Robust wav2vec 2.0 \cite{hsu2021robust},\footnote{\url{https://huggingface.co/facebook/wav2vec2-large-robust-ft-swbd-300h}} to simply transcribe all speech regions detected by the Silero VAD system,\footnote{\mbox{\url{https://github.com/snakers4/silero-vad}}} and generated an .eaf file for ELAN.\footnote{\url{https://archive.mpi.nl/tla/elan}}
Using ELAN, three annotators (2 recordings per annotator) then erased the spurious text for the Muruwari utterances (i.e.~for SLI, we simply used blank annotations to denote Muruwari regions, given the orthography is still in development) and manually corrected the English transcriptions for ASR (i.e.~for SLI, any non-blank region with text was considered English).
While the machine-generated annotations for the training and evaluation data were human-corrected, we have yet to establish inter-annotator agreement or conduct error analyses.

When correcting the English transcriptions, speech was transcribed verbatim with no punctuation except for apostrophes, i.e.~including false starts (e.g.~\textit{we we don't say}) and hesitations (e.g.~\textit{and uh it means steal}).
To facilitate searches, transcriptions were made in lower-case with the exception of proper nouns (e.g.~\textit{uh the Ngiyaamba has it uh}) and words that were spelled out by Jimmie (e.g.~\textit{you've got B U at the end of a word}).
For ASR training, the transcriptions were automatically converted to all upper-case to normalise the text to a 27-character vocabulary (26 upper-case letters + apostrophe) that matches vocabulary with which the wav2vec 2.0 Robust model was originally trained.
As we report in Appendix \ref{sec:appendix}, not re-using the original vocabulary required significantly more fine-tuning data to achieve the same performance.




Based on the corrected annotations, we extracted the speech regions into individual 16-bit 16 kHz .wav files and all the transcriptions for the English utterances into a single tab-delimited file.
A summary of the data used in this paper is given below in Table \ref{tab:recs}.
Overall, the yielded speech content contained more English than Muruwari (78\% English by duration or 66\% by number of utterances), reflecting the relatively more numerous and longer nature of the meta-linguistic commentary in English compared to the Muruwari words and phrases being commented upon.

\begin{table}[ht]
\begin{center}
\begin{tabular}{|c|cc|}
\hline
\multirow{2}{*}{\makecell{\textbf{Recording ID} \\{\small(Running time, mins)}}} & \multicolumn{2}{c|}{\bf Speech (mins)} \\
& eng & zmu \\ \hline
33-2162B (65) & 23.2  & 2.06 \\
31-1919A (65) & 16.3 & 6.28 \\
25-1581B (65) & 15.5 & 4.75 \\
25-1581A (65) & 12.1 & 4.34 \\
28-1706B (64) & 7.00  & 2.06 \\
25-1582A (35) & 6.92 & 2.68 \\ \hline
\multicolumn{1}{r}{\makecell[l]{\textbf{Total}: 5.98 hours \\ \hphantom{\textbf{Total}:} 4864 utts.}} & \makecell{81.0 mins\\3243 utts.} & \multicolumn{1}{c}{\makecell{22.2 mins\\1621 utts.}} \\
\end{tabular}
\end{center}
\vspace{-0.5em}
\caption{\label{tab:recs} Duration and number of utterances (utts.) of English and Muruwari speech yielded from labelling 6 archival recordings}
\end{table}

Notably, only a third of the total running time of the recordings was found to be speech content on average, with frequent inter- and intra-phrase pauses arising from the semi-improvised linguistic self-elicitation being undertaken by Jimmie.
A consequence of these pauses is that the VAD system segments Jimmie's speech into sequences of sentence fragments, e.g. \emph{This word...}, \emph{This word means soft...}, \emph{And also softly}.
We will return to these data characteristics in our discussion of the timed annotation tasks §\ref{sec:timed}.

Finally, we note that having had few prior experimentally-informed estimates of the minimum amounts of data required, we chose to label for our initial implementation of this workflow this specific set of 6 recordings in accordance with other project priorities.
While our deployed models are those trained on all the data, we opted to run detailed analyses on how much of the labelled data was actually necessary for adapting the SLI and ASR models to help establish estimates regarding the minimum amounts of labelled data needed to apply this workflow in other settings, and timed the annotation tasks using models trained on these minimum amounts of data.

\section{Spoken Language Identification}
\label{sec:sli}

We are interested in finding the minimum amount of training utterances required to obtain a performant system for same-speaker SLI.
As training a system with very few utterances can lead to a large variance in its performance on unseen utterances, we were particularly interested in determining the training set size at which the variance was functionally equivalent to training on all available data.

\subsection{Method}

For our SLI experiments, we first extracted speech representations from each of the 4864 English and Muruwari utterances using the SpeechBrain toolkit \cite{ravanelli2021speechbrain}, which includes a state-of-the-art SLI model trained on 107 languages of the VoxLingua107 dataset \cite{valk2021voxlingua107}.\footnote{While the model was trained to identify English (dialects unspecified), we found that the included, off-the-shelf classifier could not consistently identify Jimmie’s Australian English utterances, which were most frequently classified as Welsh (497/3243: 15.3\%) or English (321/3243: 9.8\%).}
We then performed 5000 iterations of training and evaluating logistic regression classifiers.
At each iteration, the dataset was shuffled and 20\% of the data (972 utterances) was held out as the test set.
The remaining 80\% of data (3892 utterances) was designated as the `All' training set and from which we sampled 5 additional subsets (1, 5, 10, 25, and 50 utterances per language).
We trained separate logistic regression classifiers using each of the 6 datasets (5 subsets + All), and then measured SLI performance of each classifier on the same test set using the F1 score.\footnote{Ranging between 0 (worst) and 1 (best), the F1 score is a measure of a classification system’s accuracy, taking both false positives and false negatives into account.}
Finally, we also calculated the differences between the F1 scores for the classifier trained on all the training data and each of those trained on the smaller datasets (All vs. 1, All vs. 5, All vs. 10, All vs. 25, All vs. 50).

\begin{figure}[t]
  \centering
  \includegraphics[width=0.85\linewidth]{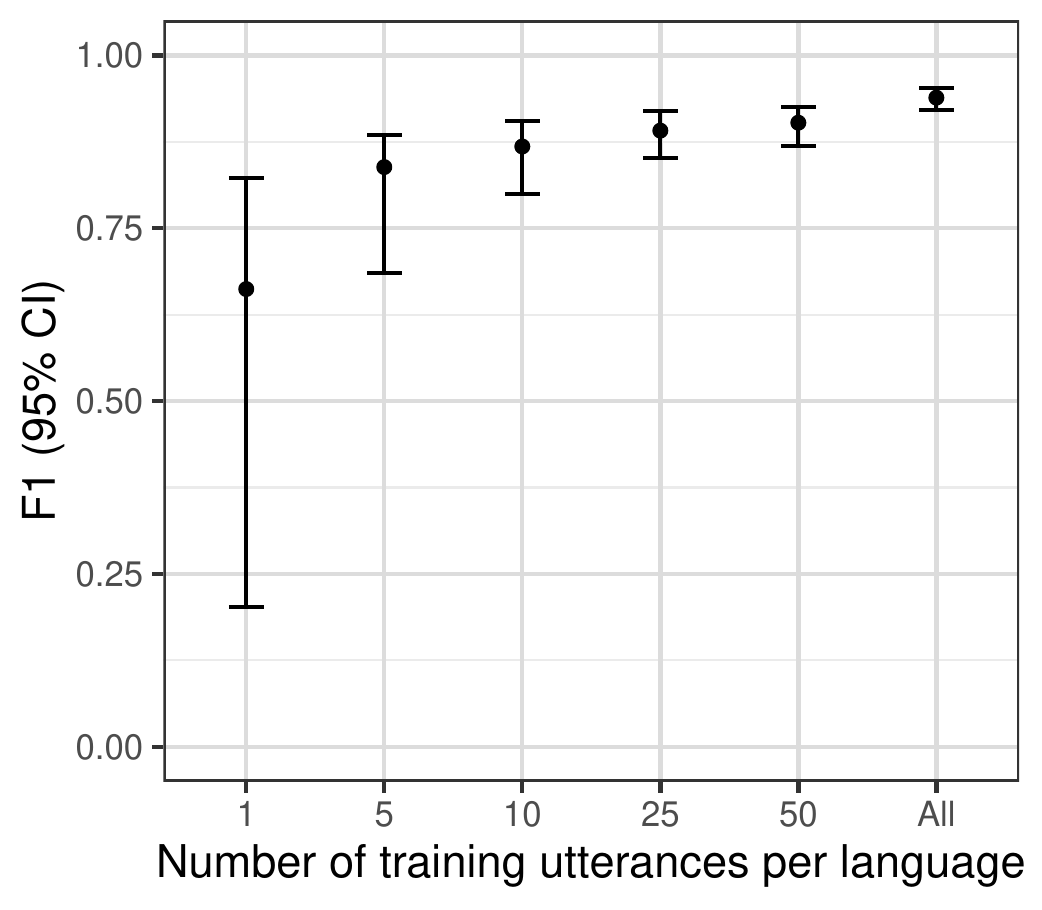}
  \caption{Two-way spoken language identification performance (English vs.~Muruwari) using logistic regression classifiers trained on SpeechBrain SLI embeddings \cite{ravanelli2021speechbrain} using varying dataset sizes (1, 5, 10, 25, 50 utterances per language, and All available data: 3892 utterances). Points represent mean F1 and error bars the 95\% bootstrap confidences intervals over 5000 iterations.}
  \label{fig:sli-results}
\end{figure}

\subsection{Results}

Figure \ref{fig:sli-results} displays the mean F1 scores for each of the training dataset sizes.
The error bars represent the 95\% bootstrap confidence interval (CI) for the mean obtained over 5000 iterations.
Using all the training data resulted in the highest SLI performance of 0.93 [95\% CI: 0.91, 0.95].
Of the smaller dataset sizes, the 50-, 25-, and 10-utterance training subsets performed similarly with mean F1 scores of 0.90 [95\% CI: 0.87, 0.93], 0.89 [95\% CI: 0.85, 0.92], and 0.87 [95\% CI: 0.79, 0.91], respectively.
The smallest two dataset sizes showed yet lower SLI performance with mean F1 scores for 5 utterances at 0.84 [95\% CI: 0.69, 0.89] and 1 utterance at 0.66 [95\% CI: 0.20, 0.82].

\begin{table}[t]
\begin{center}
\begin{tabular}{|l|c|c|}
\hline
\textbf{Comparison} & \makecell{\textbf{Difference in F1}\\Mean, {[}95\% CI{]}: CI width} \\ \hline
a. All vs. 1   & 0.28, [0.11, 0.74]: 0.63\\
b. All vs. 5   & 0.10, [0.05, 0.25]: 0.20 \\
c. All vs. 10  & 0.07, [0.03, 0.14]: 0.11  \\
d. All vs. 25  & 0.05, [0.02, 0.09]: 0.07 \\ 
e. All vs. 50  & 0.04, [0.01, 0.07]: 0.06  \\ \hline
\end{tabular}
\end{center}
\caption{\label{tab:sli-confs} Mean difference in F1 and 95\% bootstrap confidence intervals (lower and upper bounds, and width) for the difference in means for the performance on a spoken language identification task using logistic regression classifiers trained of varying dataset sizes (1, 5, 10, 25, 50 utterances per language, and All available training data: 3892 utterances)}
\end{table}

Table~\ref{tab:sli-confs} displays the mean differences and the corresponding confidence intervals for the mean differences in F1 scores for the classifier trained on all the training data (All) and each of those trained on the smaller datasets (1, 5, 10, 25, 50 utterances per language).
On average, using only 1 utterance of English and Muruwari results in a system that is 28 percentage points worse than using all the data (Table~\ref{tab:sli-confs} a).
While using 5 or 10 utterances resulted in similar average differences compared to using all the data (10 vs 7 percentage points, respectively), the difference is nearly twice as variable when only 5 utterances per language are used (CI width: 20 percentage points). 

Answering our SLI-related questions, then: 1) using 10 utterances per language yields systems whose average performance is within 10 percentage points of using all the data (3892 utterances). 2) a logistic regression classifier suffices for two-way same-speaker SLI using off-the-shelf speech embeddings for SLI \cite[][]{ravanelli2021speechbrain}.

\section{Automatic Speech Recognition}
\label{sec:asr}

Recall that for ASR, we seek to answer the following questions: 1) What amount of transcribed speech is sufficient to reliably achieve better than off-the-shelf performance for transcribing Jimmie's Australian English? 2) Using the same amount of transcribed speech, to what extent can ASR system performance be further increased when supplemented with a language model trained on external texts?
In this section, we report on experiments conducted in order to answer these questions.

\subsection{Method}

In all our fine-tuning experiments, we fine-tuned the Robust wav2vec 2.0 model over 50 epochs, evaluating every 5 epochs (with an early-stopping patience of 3 evaluations).
All training runs started from the same off-the-shelf checkpoint and we kept constant the training hyperparameters, all of which can be inspected in the model training script on GitHub.
We varied as the independent variable the amount and samples of data used to fine-tune the model and measured as the dependent variable the word error rate (WER).\footnote{Ranging from 0\% (best) to 100\% (worst), word error rate (WER) is a measure of the accuracy of an ASR system, taking
into account substitutions (wrongly predicted words), additions (erroneous extra words) and deletions (missing words).} 

In all our experiments, we split the total 81 minutes of transcribed English speech into an 80\% training set (65 minutes) and a 20\% testing set (16 minutes).
The training split of 65 minutes was designated as the 100\% training set from which we sampled smaller subsets consisting of 52 minutes (80\% of training split), 39 minutes (60\% of training split), 26 minutes (40\% of training split), 13 minutes (20\% of training split), 6.5 minutes (10\% of training split), 3.25 minutes (5\% of training split), and 0.65 minutes (1\% of training split).

We fine-tuned 8 separate models with varying amounts of data and evaluated their performance on the same test set to obtain a first estimate of an amount of data sufficient to achieve better than off-the-shelf performance. 
We then created 10 new 80/20 training/testing splits for cross-validation in order to establish the variability in WER when only using that minimal amount of data.

We were also interested in whether supplementing the ASR system with a language model further reduced the WER.
Our initial labelling work revealed that many errors made by the off-the-shelf system were particularly related to domain- and region-specific English words (e.g. \textit{spear}, \textit{kangaroo}).
With permission from the maintainers of the Warlpiri-to-English dictionary, we extracted 8359 English translations from example sentences to obtain in-domain/-region sentences in English, e.g.~\textit{The two brothers speared the kangaroo}.

We used this data to train a word-level bigram model using KenLM \cite{heafield2011kenlm}.
While we opted to extract sentences from the Warlpiri-to-English dictionary given it is the largest of its kind for an Australian language, this corpus of sentences still only amounts to 75,425 words (4,377 unique forms), and as such we opted for a bigram model over a more conventional 3- or 4-gram model.
With the only change being the inclusion of the language model, we then fine-tuned 10 additional models using the same training and testing splits.

\begin{table}[t]
\vspace{1em}
\begin{center}
\begin{tabular}{|l|c|c|}
\hline

\textbf{Training set size} & \textbf{WER} & \textbf{CER} \\ \hline

\makecell[l]{
a. 65 minutes (100\%)} & 10.1\% & 4.2\% \\ 
b. 52 minutes (80\%)  & 10.1\% & 4.4\%\\
c. 39 minutes (60\%)  & 11.8\% & 5.2\% \\
d. 26 minutes (40\%)  & 12.3\% & 5.5\% \\ 
e. 13 minutes (20\%)  & 13.2\% & 6.1\% \\ 
f. 6.5 minutes (10\%)  & 13.4\% & 6.1\% \\ 
g. 3.25 minutes (5\%)  & 15.1\% & 6.7\% \\ 
h. 0.65 minutes (1\%)  & 19.1\% & 8.8\% \\ 
\hline

i. \makecell[l]{Off-the-shelf (0\%)} & 36.3\% & 21.5\% \\ \hline

\end{tabular}
\end{center}
\vspace{-0.5em}
\caption{\label{tab:w2v2-datasets} Word error rates (WERs) achieved from fine-tuning the same wav2vec 2.0 model (large-robust-ft-swbd-300h) over 50 epochs using various subsets of data from 65 minutes of Australian English archival audio data.}
\end{table}

\subsection{Results}

Table~\ref{tab:w2v2-datasets} displays the word error rates (WERs) achieved by a Robust wav2vec 2.0 model fine-tuned with various amounts of transcribed speech. 
The baseline WER achieved by the off-the-shelf model with no additional fine-tuning is 36.3\%.
Training with all 65 minutes of data yielded a topline WER of 10.1\%.
Remarkably, training with less than 1 minute of speech was sufficient to decrease the WER to 19.1\%.
As a first estimate, the amount of training data that sufficiently improves on the off-the-shelf model appears to be 0.65 minutes of transcribed speech. 

\begin{figure}[t]
  \centering
  \includegraphics[width=0.85\linewidth]{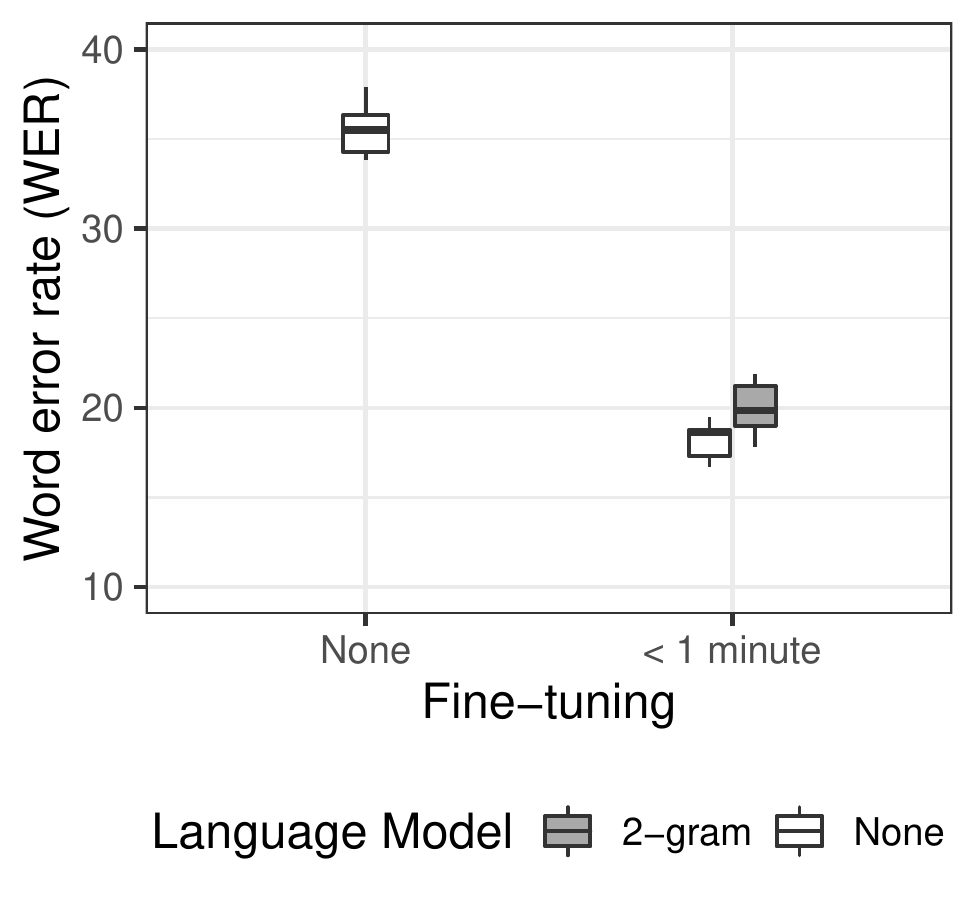}
  \vspace{-1em}
  \caption{Variability in word error rates of training and testing Robust wav2vec 2.0 models over 10 iterations using different samples in the training and testing datasets, holding constant the size of the training set (1\% of training set = 0.65 minutes or 39 seconds, on average) and testing set (16 minutes). The off-the-shelf model without fine-tuning was also evaluated on the same 10 testing sets.}
  \label{fig:asr-exp3}
  \vspace{-1em}
\end{figure}

\begin{table*}[!b]
\begin{center}
\begin{tabular}{|c|cccccc|}
\multicolumn{1}{r}{} & \multicolumn{5}{c}{\makecell{Time taken in minutes (Annotator)}} \\
\hline
\multirow{3}{*}{\makecell[c]{\textbf{Recording ID}\\\ (Running time, mins)}} & \multicolumn{2}{c}{\textbf{Segmentation only}} & \multicolumn{4}{|c|}{\makecell{\textbf{Transcription only}}} \\

& \multirow{2}{*}{Manual} & \multirow{2}{*}{\makecell{Assisted\\VAD+SLI}} & \multicolumn{1}{|c}{\multirow{2}{*}{Manual}} & \multicolumn{3}{c|}{\makecell{Assisted: ASR systems, A--C}} \\
& & & \multicolumn{1}{|c}{} & A & B & C \\ \hline
33-2171A/S1 (31) & {88} (A1) & 81 (A2) & \multicolumn{1}{|c}{-} & - & 54 (A4)  & 53 (A3) \\ 
33-2163A/S1 (33) & 83 (A2) & 84 (A1) & \multicolumn{1}{|c}{-} & - & 57 (A3)  & 66 (A4) \\ \hline
33-2167B/S2 (32) & - & - & \multicolumn{1}{|c}{\makecell{96/87\\(A1/A2)}} & \makecell{55/71\\(A3/A4)} & - & - \\ \hline
\multicolumn{1}{r}{\textbf{Mean time taken, in minutes}} & {\textbf{85.5}} & {\textbf{82.5}} & {\textbf{91.5}} & {\textbf{63.0}} & \textbf{55.5} & \multicolumn{1}{c}{\textbf{59.5}} \\

\end{tabular}
\end{center}
\vspace{-0.5em}
\caption{\label{tab:workflow} Time taken to annotate recordings by four annotators (A1--A4) with and without machine assistance. In the segmentation task, annotators corrected the segmentations by the voice activity detection (VAD) and spoken language identification systems (SLI: trained on 10 utterances per language), or they manually annotated speech regions. In the transcription task, annotators were given intervals of English speech without any accompanying text (manual transcription), or text generated by one of three ASR (A, B C) systems differing in accuracy.
System A was an off-the-shelf Robust wav2vec 2.0 model \cite{hsu2021robust} with no fine-tuning (word error rate/character error rate: 36/22).
Systems B (19/7) and C (14/6) were Robust wav2vec 2.0 models fine-tuned on 39 minutes of transcribed English speech, and System C supplemented with a bigram language model trained on external texts.}
\vspace{-0.5em}
\end{table*}

To verify that fine-tuning with only 1\% of our training data does consistently yield a better than off-the-shelf WER, we conducted cross-validation experiments using 10 additional 80/20 training/testing splits, each time using only 1\% of the data from the training split (0.65 minutes or 39 seconds on average).

Figure~\ref{fig:asr-exp3} displays the results of our cross-validation experiments.
First, evaluating the off-the-shelf model on the 10 test sets, we found the baseline mean WER to be 35.6\% (standard deviation, SD: 1.48\%; range: 33.8--37.9\%).
The mean WER of the models fine-tuned with only 1\% of data and without a language model was found to be 18.2\% (SD: 0.99\%; range: 16.7--19.5\%).
These results demonstrate that fine-tuning with less than 1 minute of speech consistently yields better than off-the-shelf performance.

When a bigram language model was used for decoding, we found that the mean WER increased to 20.0\% (SD: 1.48\%; range: 17.8--21.9\%) for the fine-tuned models.
These results are inconsistent with our earlier experiments (reported in Appendix \ref{sec:appendix}), where we fine-tuned the same off-the-shelf model with 39 minutes of data.
In these experiments, decoding with the same bigram model did lead to WER improvements, suggesting that more careful calibration and weighting of the language model may be required in near-zero shot adaptation scenarios.

To answer our ASR-related questions, then: 1) 39 seconds on average of speech on average is sufficient to achieve a better than off-the-shelf performance for transcribing Jimmie’s Australian English speech.
2) the effect on ASR performance of a language model is inconclusive (cf.~Appendix \ref{sec:appendix}).

\section{Timed annotation tasks}
\label{sec:timed}

In addition to helping provide estimates of the contents of recordings for review by an authorised person, another purpose of this workflow is to help reduce the time required to annotate speech in such a way that excerpts from cleared recordings can be easily extracted for use in relevant initiatives, e.g. creating language learning materials.

\begin{figure}[H]
  \centering
  \vspace{0.5em}
  \includegraphics[width=0.65\linewidth]{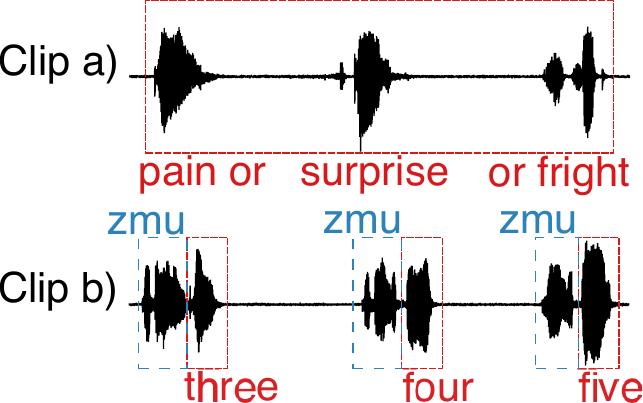}
  \vspace{0.5em}
  \caption{Desired annotations for two excerpts of speech from the Jimmie Barker recordings. Clip a) shows a sequence of sentence fragments in English, to be annotated as a single utterance. Clip b) shows alternating Muruwari (zmu) and English speech, to be annotated as 6 utterances.}
  \label{fig:frags}
\end{figure}

The initial process of annotating speech for this purpose involves two tasks: segmentation and transcription, which we illustrate in Figure~\ref{fig:frags} using two clips of Jimmie's speech.
In segmentation, the annotator identifies regions of speech and non-speech and also which of the speech regions is English or Muruwari.
For a sequence of English sentence fragments such as those in Clip a), the utterances can simply be merged into one.
For mixed-language regions such as those in Clip b), separate utterances should be created to allow the Muruwari speech to be easily extracted for use in language learning materials.
To create transcriptions for indexing, the annotator transcribes the English segments, given regions segmented and identified as English.
We conducted a set of timed annotation tasks to evaluate to what extent the machine-assisted workflow reduces the time taken to perform these two tasks.

As detailed in Table~\ref{tab:workflow}, we gathered for our timed annotation tasks three different recordings approximately 30 minutes in length that were not part of the training and evaluation recordings in the previous experiments.
For each timed task, annotators were asked to perform only segmentation or only transcription.
For segmentation, they either manually created all time boundaries or corrected machine-derived ones from the VAD and SLI systems.
For transcription, they either manually typed in the transcriptions for English speech or corrected machine-derived ones from an ASR system.
We tested ASR systems developed earlier in our research (reported in Appendix \ref{sec:appendix}), that was fine-tuned on 39 minutes of Jimmy's Australian English speech, and reached a WER/CER of 19/7, as well as a version of the same system augmented with a bigram language model which reached a WER/CER of 14/6.
The three recordings and the four annotators and the six annotation tasks were counter-balanced such that each annotator listened to each recording for a given task exactly once.

The segmentation task took 85.5 minutes of work for a 30-minute recording without machine assistance and 82.5 minutes when assisted.
That is, correcting time boundaries, inserting missing intervals or removing erroneous ones, and merging/splitting machine-derived segmentations takes nearly the same amount of time as placing these boundaries manually. 
The waveforms in Figure~\ref{fig:frags} illustrate how the acoustics of alternating Muruwari and English separated by brief pauses look indistinguishable from English sentence fragments separated by similar amounts of pauses --- leading to sub-optimal segmentations using a standard, sequential VAD-then-SLI pipeline.
The mixed-language nature of this speech may require jointly optimising the VAD and SLI steps.

The transcription task took 91.5 minutes of work for a 30-minute recording without machine assistance and on average 59.3 minutes when assisted (a 35\% reduction).
We found no meaningful difference between the correction times for transcriptions generated by ASR systems with different levels of accuracy.
For transcriptions produced by an off-the-shelf system (WER/CER: 36/22), the correction time was 63 minutes.
For systems fine-tuned with 39 minutes of transcribed speech, WER/CER: 19/7 and 14/6, the correction times were 55.5 and 59.5 minutes, respectively.

The closeness in transcription correction times may relate to how an English ASR system whose WER is 30\% or less produces good enough transcriptions for editing, according to a crowd-sourced study \cite{gaur2016effects}.
Here, our transcribers' tolerance for the relatively less accurate off-the-shelf system (WER 36\%) may be attributable to their familiarity with the speech domain and speaker \citep{Sperber2017b}, having collectively spent nearly 40 hours correcting transcriptions of Jimmie's English by the time we conducted the timed tasks.
These results suggest that, where correction is permissible by L1-speaking transcribers of the metalanguage, the time savings over manual transcription could still be gained using an off-the-shelf system that achieves a WER of 30--36\% or less for the metalanguage in the recordings.

Nevertheless, we find that the machine-assisted workflow does offer time savings over a fully manual workflow \citep[in line with previous work, e.g.:][]{sperber-etal-2016-optimizing,Sperber2017b}.
Specifically, we find that the machine-assisted workflow offers a \PCdiff{} reduction in overall time to identify regions in the target language and metalanguage and also transcribe the latter, requiring 2.36 hours (82.5 + 59.3 mins) of correction time for a 30-minute recording compared to a fully-manual one which requires 2.95 hours (85.5 + 91.5 mins).
Unlike the manual workflow, the fully-automatable workflow can derive first-pass transcriptions to help an authorised person triage recordings.

\section{Towards a Muruwari orthography}
\label{sec:doc}

As mentioned above, the Muruwari orthography is still currently in development.
In this section, we provide a brief overview of how transcriptions of the English metalanguage are being used to aid in the development of the Muruwari orthography.

A key source of information on Muruwari phonemes and words of interest to the current Muruwari community are two 1969 recordings in which Jimmie Barker discusses an early Muruwari wordlist \citep{mathews1902}. 
This wordlist was created by linguist R.H. Mathews and consists of Muruwari words in his romanisation along with English translations.
Using this wordlist, the documentation team is able to shortlist Muruwari words whose romanisation is suggestive of containing sounds of interest (e.g. dental consonants), and then quickly locate in these recordings Jimmie's pronunciation of the words and associated commentary using the time-aligned English transcripts generated for the two recordings.
Here, the English transcripts provide significantly more streamlined access to untranscribed Muruwari utterances than browsing the recordings in real time.
Once verified of containing the sounds of interest, the documentation team is able to extract snippets of these words to be included in the community consultation process.

\section{Conclusion}
\label{sec:conc}

Many hours of unannotated speech from endangered languages remain in language archives and inaccessible to community members and language learning programs.
The time-intensive nature of annotating speech creates one bottleneck, with an additional one occurring for speech in restricted access corpora that authorised community members must vet before annotation can begin.
For a particular genre of recordings where speech in the endangered language is intermixed with a metalanguage in a more widely-used language such as English, we proposed a privacy-preserving workflow using automated speech processing systems to help alleviate these bottlenecks.

The workflow leverages voice activity detection (VAD) to identify regions of speech in a recording, and then spoken language identification (SLI) to isolate speech regions in the metalanguage and transcribes them using automatic speech recognition (ASR).
The uncorrected transcriptions provide an estimate of the contents of a recording for an authorised person to make initial decisions on whether it can be listened to by those with lower levels of access to correct the transcriptions, which, collectively, help index the corpus.
This workflow can be implemented using a limited amount of labelled data: 10 utterances per language for SLI and 39 seconds of transcribed speech in the metalanguage for ASR.
The workflow reduces metalanguage transcription time by \PCdiff{} over manual transcription and similar time savings may be achievable with an off-the-shelf ASR system with a word error rate of 36\% or less for the metalanguage in the target recordings.

Given our use case, the present demonstration of the workflow was limited to the scenario of processing single-speaker monologues with a mix of Muruwari and English, the latter of which made possible the use of a state-of-the-art model trained for English ASR \cite[Robust wav2vec 2.0: ][]{hsu2021robust} and also for transcriptions to be corrected by first language speakers of English.
Our work also revealed that VAD and SLI systems require further optimisation for mixed-language speech.

We hope our demonstration encourages further experimentation with model adaptation with limited data for related use cases. 
For dialogues between a linguist and language consultant, for example, speaker diarisation could be added via few-shot classification using speech representations for speaker recognition \cite[e.g. SpeechBrain SR embeddings: ][]{ravanelli2021speechbrain}.
With user-friendly interfaces like Elpis \cite{foley2018building}, for which wav2vec 2.0 integration is underway (Foley, pers. comm.), we hope to see more streamlined access to pre-trained models for language documentation workflows and, consequently, more streamlined access to the recorded speech for community members and language learning programs.


\bibliography{computel3}
\bibliographystyle{acl_natbib_nourl}

\appendix

\section{Fine-tuning with a re-initialised vocabulary}
\label{sec:appendix}

In this section, we describe an earlier set of ASR fine-tuning experiments which were analogous to those reported in §\ref{sec:asr}, except for the manner in which vocabulary (i.e. character set) was configured.
Following recommended fine-tuning practice,\footnote{\url{https://huggingface.co/blog/fine-tune-wav2vec2-english}} we initialised a linear layer whose output size corresponds to set of characters to be predicted (e.g. `A', `B', ...) and is derived from the target training dataset.
However, this guidance presupposes that the pre-trained model being fine-tuned is one with no prior fine-tuning for ASR on the same language.

Given the size of our available training data (total 65 minutes), we chose to continue to train the Robust wav2vec 2.0 model,\footnote{\url{https://huggingface.co/facebook/wav2vec2-large-robust-ft-swbd-300h}} already fine-tuned for English ASR on 300 hours of Switchboard \cite{godfrey1992switchboard}. 
The results of fine-tuning this model using various-sized subsets of our training data is reported below in Table \ref{tab:w2v2-datasets-old}.
Notably, fine-tuning with only 13 minutes of data resulted in a significantly worse than off-the-shelf performance (98\% vs. 37\%, off the shelf).
By deriving labels for the linear layer from our training dataset, the label mappings were scrambled (e.g. from Output 4 = `E' to Output 4 = `C'), yielding gibberish predictions during initial fine-tuning.
Through this fine-tuning process, 39 minutes of training data were required for the model to (re-)learn the appropriate parameters for English ASR.

\begin{table}[t]
\begin{center}
\begin{tabular}{|l|c|c|}
\hline

\textbf{Training set size} & \textbf{WER} & \textbf{CER} \\ \hline

\makecell[l]{
a. 65 minutes (100\%)} & 11\% & 5\% \\ 
b. 52 minutes (80\%)  & 13\% & 5\%\\
c. 39 minutes (60\%)  & 16\% & 6\% \\
d. 26 minutes (40\%)  & 37\% & 14\% \\ 
e. 13 minutes (20\%)  & 98\% & 78\% \\ 
\hline

f. \makecell[l]{Off-the-shelf (0\%)} & 37\% & 22\% \\ \hline

\end{tabular}
\end{center}
\vspace{-0.5em}
\caption{\label{tab:w2v2-datasets-old} Word error rates (WERs) achieved from fine-tuning the same wav2vec 2.0 model (large-robust-ft-swbd-300h) over 50 epochs using various subsets of data from 65 minutes of Australian English archival audio data.}
\end{table}

By contrast, in our experiments reported above in §\ref{sec:asr}, we adapted our datasets to match the vocabulary of the tokeniser included with the off-the-shelf model.
By doing so, we were able to achieve better than off-the-shelf ASR performance using only 39 seconds of training data.

\begin{figure}[H]
  \centering
  \includegraphics[width=0.85\linewidth]{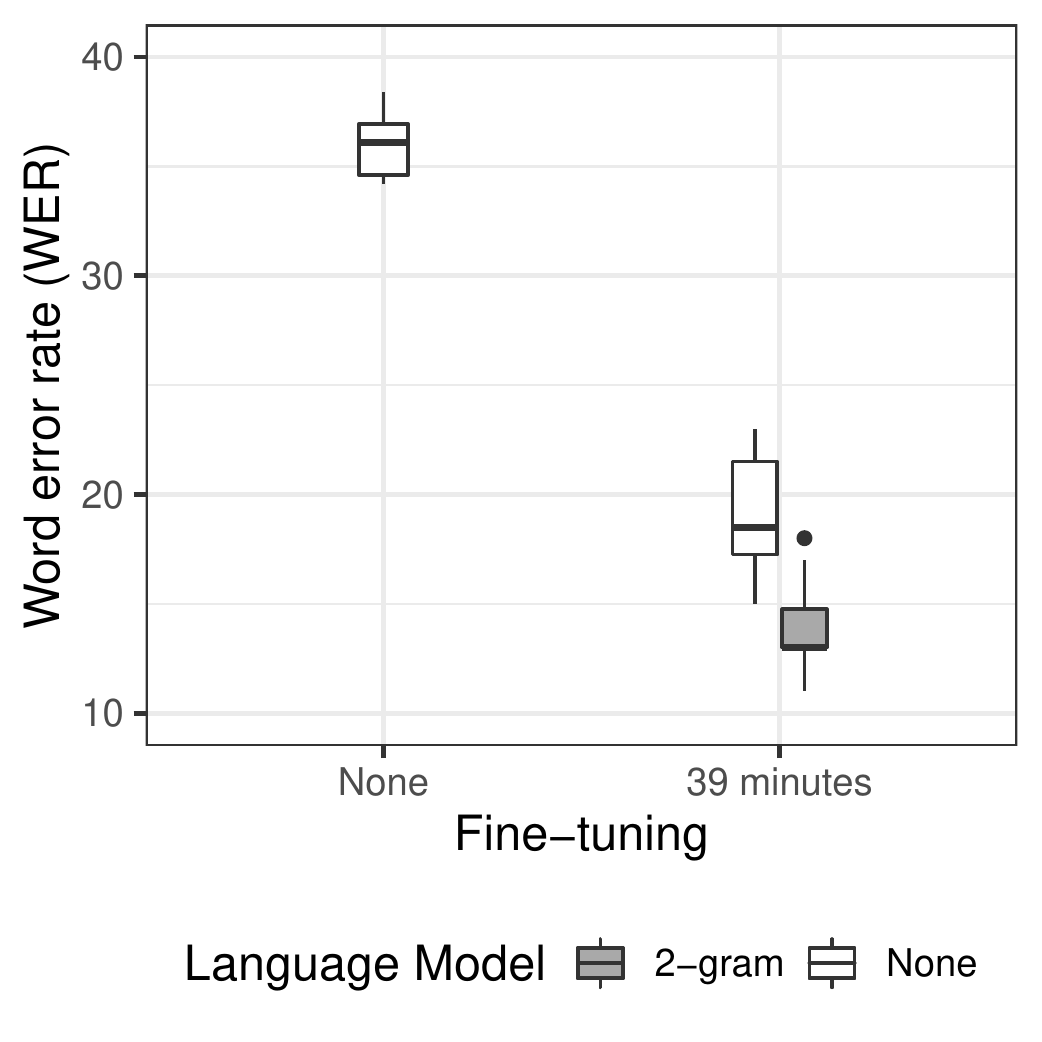}
  \vspace{-1em}
  \caption{Variability in word error rates of training and testing Robust wav2vec 2.0 models over 10 iterations using different samples in the training and testing datasets, holding constant the size of the training set (39 minutes) and testing set (16 minutes). The off-the-shelf model without fine-tuning was also evaluated on the same 10 testing sets.}
  \label{fig:asr-exp3-old}
  \vspace{-1em}
\end{figure}

Yet, unlike those experiments reported above, the addition of a language model to models fine-tuned with a re-initialised vocabulary yielded better performance. 
As shown in Figure \ref{fig:asr-exp3-old}, the mean WER of the models fine-tuned with 39 minutes of data and without a language model was found to be 19.5\% (SD: 2.98\%; range: 15--23\%).
When a bigram language model was included, we found that the mean WER decreased to 14\% (SD: 2.30\%; range: 11--18\%).
These findings suggest that while the addition of a language model can be beneficial more experimentation is needed to inform best practices for calibrating and/or weighting the language model in near-zero shot learning scenarios.

\end{document}